\documentclass[a4paper]{article}

\usepackage[english]{babel}
\usepackage[utf8x]{inputenc}
\usepackage[T1]{fontenc}

\usepackage[a4paper,top=3cm,bottom=2cm,left=3cm,right=3cm,marginparwidth=1.75cm]{geometry}

\usepackage{amsmath}
\usepackage{graphicx}
\usepackage[colorinlistoftodos]{todonotes}
\usepackage[colorlinks=true, allcolors=blue]{hyperref}

\usepackage{subcaption}
\usepackage{graphicx}
\usepackage{algorithm}
\usepackage[noend]{algpseudocode}

\newcommand{\refabeqp}[1]{(Eq.~\ref{eq:#1})}

\newcommand{\citejacklicbook}[0]{Jackli\v{c} et al. (2000) }

\title{Sampling Superquadric Point Clouds with Normals}
\author{Paulo Ferreira}

\begin{document}
\maketitle

\abstract{
Superquadrics provide a compact representation of common shapes and have been used both for object/surface modelling in computer graphics and as object-part representation in computer vision and robotics. Superquadrics refer to a family of shapes: here we deal with the superellipsoids and superparaboloids. Due to the strong non-linearities involved in the equations, uniform or close-to-uniform sampling is not attainable through a naive approach of direct sampling from the parametric formulation. This is specially true for more `cubic' superquadrics (with shape parameters close to $0.1$). We extend a previous solution of 2D close-to-uniform uniform sampling of superellipses to the superellipsoid (3D) case and derive our own for the superparaboloid. Additionally, we are able to provide normals for each sampled point. To the best of our knowledge, this is the first complete approach for close-to-uniform sampling of superellipsoids and superparaboloids in one single framework. We present derivations, pseudocode and qualitative and quantitative results using our code, which is available online.
}

\section{Introduction}

Superquadrics were introduced by Barr (1981) and this name usually refers to a family of shapes that includes superellipsoids, superhyperboloids and supertoroids. The \textit{super} part of the name refers to the fact that the original curve (e.g. ellipse) is exponentiated; and the \textit{oid} suffix refers to the 3D case. Thus, the superellipsoid is  the `3D version' of the exponentiated ellipse. Here we use superquadrics to mean the superellipsoids plus the superparaboloids; we do not deal with the superhyperboloids and supertoroids. The superparaboloid literature is scarcer and to the best of our knowledge there is no complete formulation of it that also relates it to the superellipsoids: here we provide our own (Sec.~\ref{sec:superparaboloids_formulation}). 

\begin{figure}[h!]
    \centering
    \includegraphics[width=0.4\linewidth]{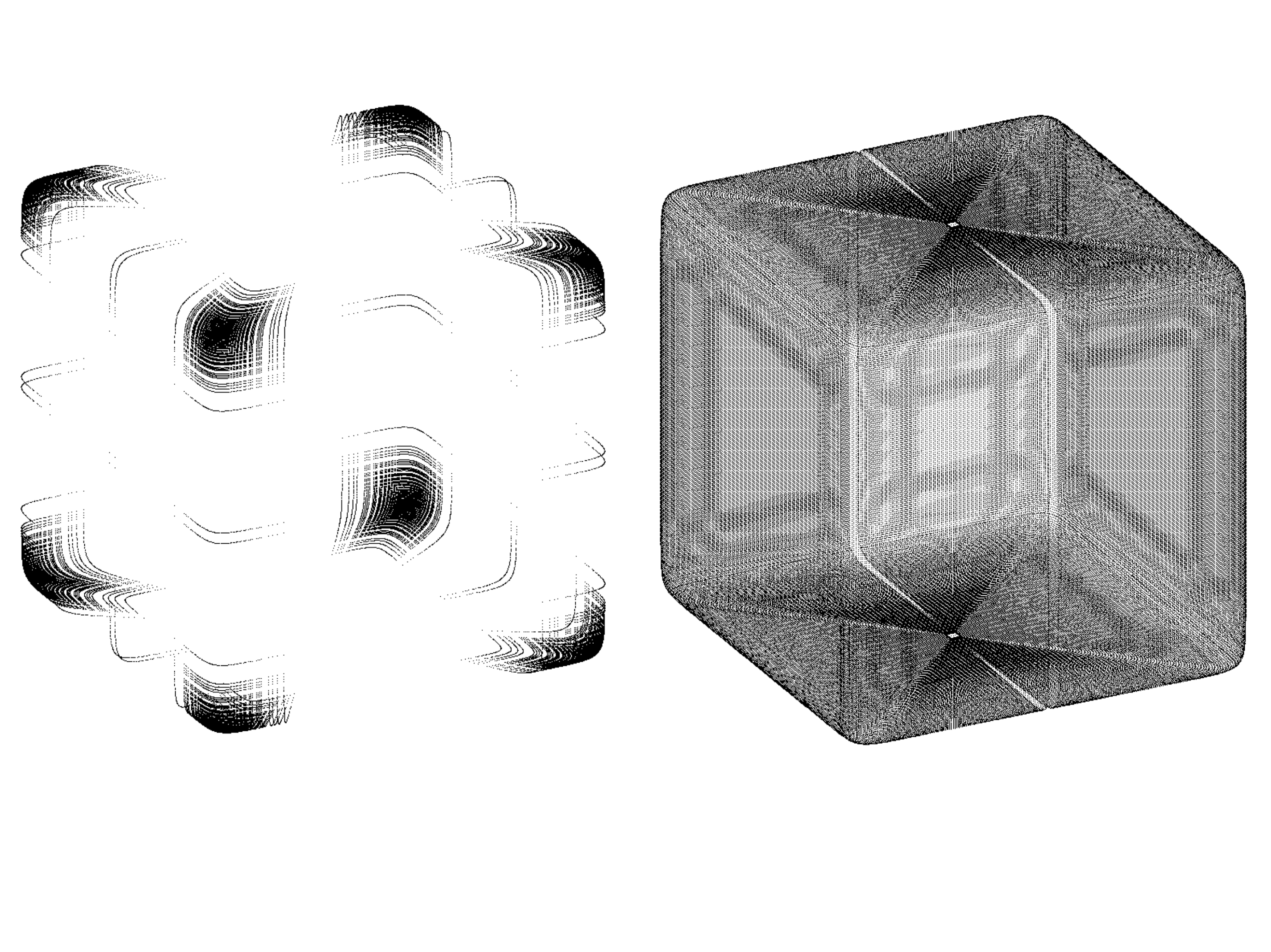} 
    \caption{Naive parametric approach (left) versus ours (right) for sampling a `cube' superquadric. Strong non-linearities lead the naive approach to sample mostly from regions of high curvature; we are able to achieve close-to-uniform results.} 
    \label{fig:naive_vs_unif_cube} 
\end{figure}

\begin{figure}[t!] 
  \begin{subfigure}[b]{0.5\linewidth}
    \centering
    \includegraphics[width=0.6\linewidth]{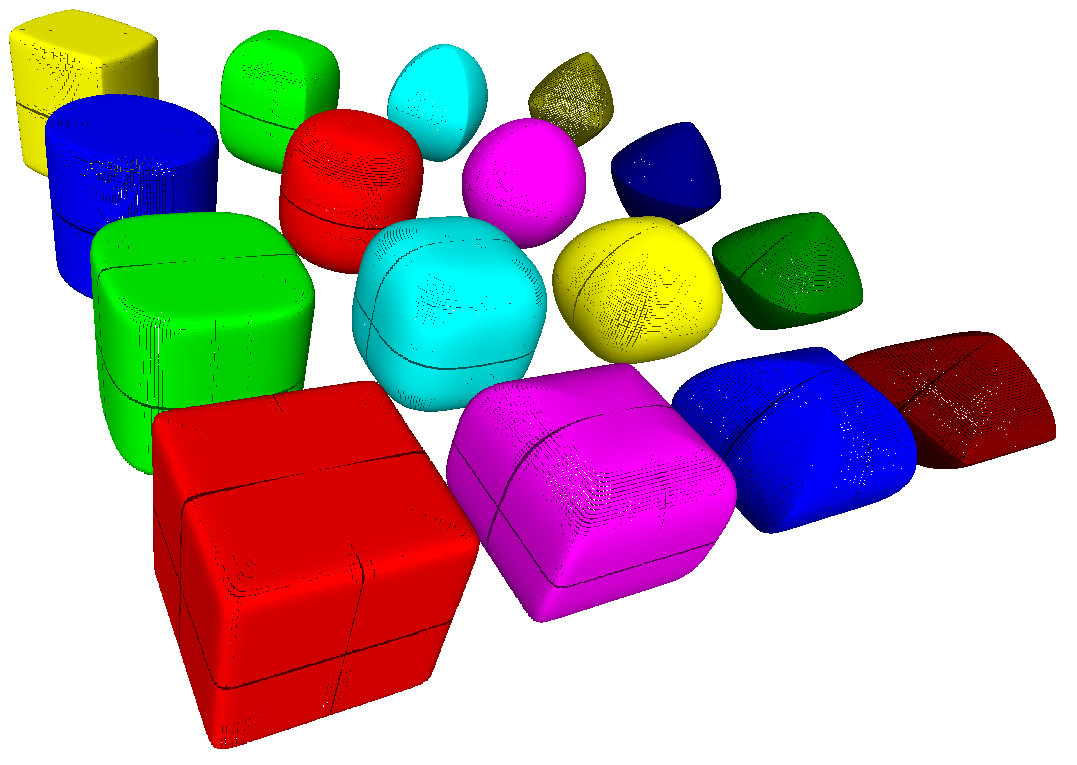} 
    \caption{Superellipsoids} 
    \label{fig:task_sim_roll} 
    \vspace{2ex}
  \end{subfigure}
  \begin{subfigure}[b]{0.5\linewidth}
    \centering
    \includegraphics[width=0.6\linewidth]{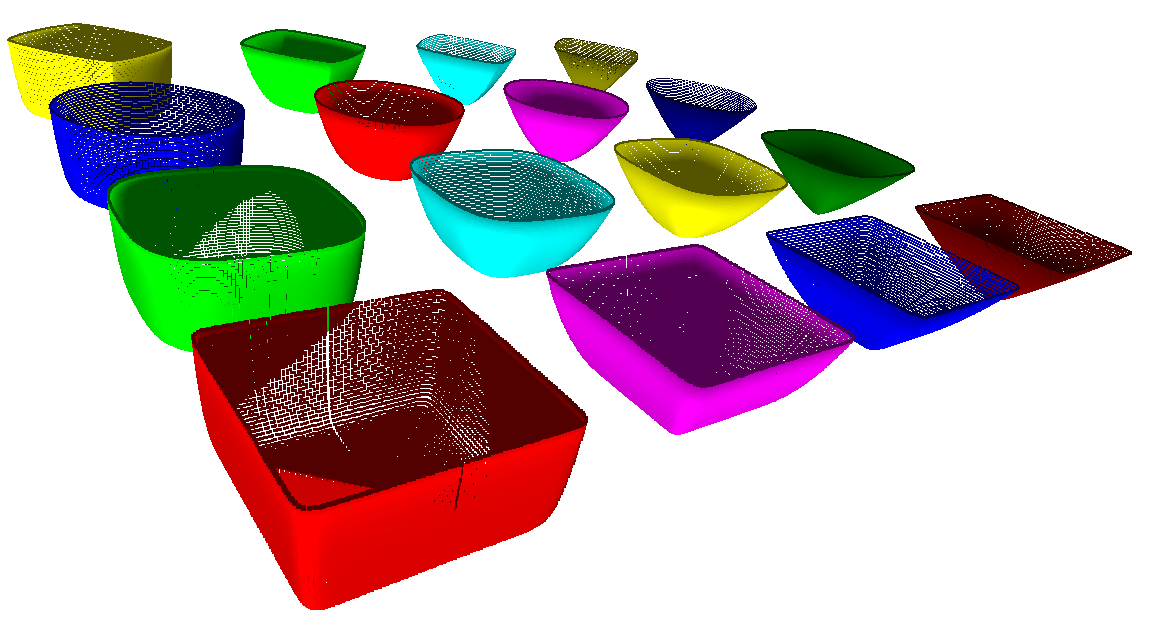} 
    \caption{Superparaboloids} 
    \label{fig:task_sim_cut} 
    \vspace{2ex}
  \end{subfigure} 
  \begin{subfigure}[b]{0.5\linewidth}
    \centering
    \includegraphics[width=0.5\linewidth]{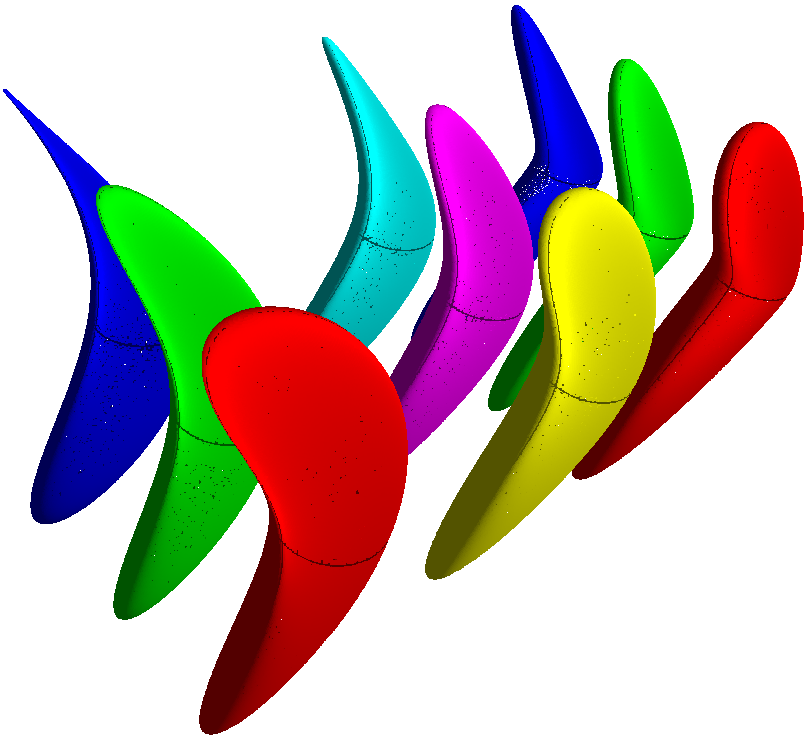} 
    \caption{Tap./bent superellipsoids} 
    \label{fig:task_sim_hammer} 
  \end{subfigure}
  \begin{subfigure}[b]{0.5\linewidth}
    \centering
    \includegraphics[width=0.6\linewidth]{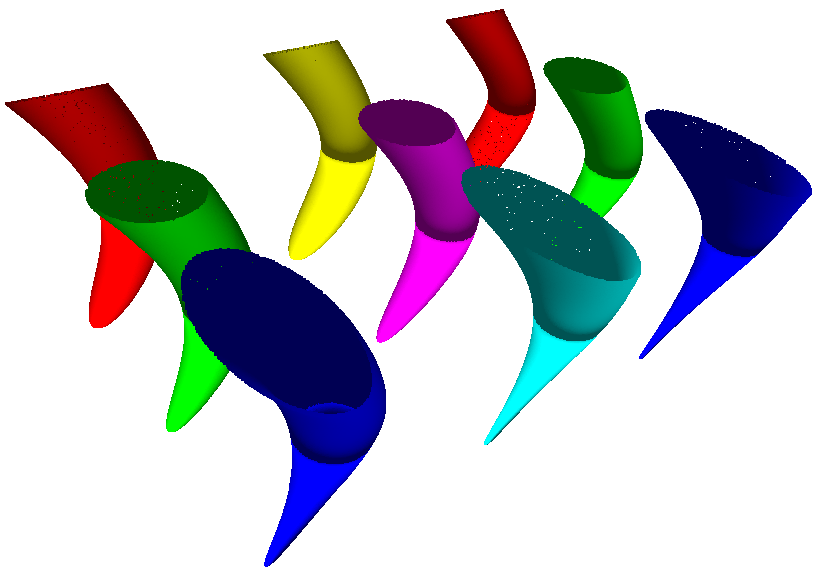} 
    \caption{Tap./bent superparabol.} 
    \label{fig:task_sim_lift} 
  \end{subfigure} 
  \vspace{0.1cm}
  \caption{Examples generated with our method: (a) superellipsoids; (b) superparaboloids; (c) tapered and bent superellipsoids; and (d) tapered and bent superparaboloids}
  \label{fig:intro_superquadrics}
\end{figure}

Close-to-uniform sampling is essential for accurate and realistic graphical modelling and rendering. Superquadrics provide a compact representation that can span a variety of shapes (Fig.~\ref{fig:intro_superquadrics}). A naive parametric approach does not provide a satisfactory output and it is specially problematic for highly cubical superquadrics \cite{Franklin1981,Pilu1995}. In Fig.~\ref{fig:naive_vs_unif_cube} we show a comparison between the naive approach and ours. All results and images in this paper were obtained using our approach and implementation in Matlab; the code is available online\footnote{You can find a demo version of the code at: https://github.com/pauloabelha/enzymes/tree/master/demos/SQ. You can run GetAllDemoSQs.m as a start to see different superquadrics being sampled.}.

Superquadrics are used in scientific visualisation \cite{Kindlmann2004}, medical image analysis \cite{Bardinet1996}, graphical modelling \cite{Talu2011,Hsu2012}, object recognition 
\cite{Andreopoulos2012,Krivic2004105,BiegelbauerICRA2007} and object segmentation/decomposition in general \cite{JacklicSQBook,Page2003} and in particular for computer vision for robotics
\cite{Varadarajan2011b,BiegelbauerICRA2007,Duncan2013,Drews2010} and grasping for robotics \cite{Uckermann2012,Varadarajan2011a,Strand2010,Guo2014,Cocias2012,Aleotti2012}. 
We refer the reader to \citejacklicbook for a thorough exposition of superquadrics.

To the best of our knowledge there is no work on uniform sampling of superquadric surfaces. Pilu and Fisher (1995) provide a method for uniformly sampling superellipses. Here, we derive a method for the 3D case of superellipsoids and we also define and derive for the superparaboloid. Regarding superparaboloids, the first time they were given a parametric formulation is in L{\"o}ffelmann and Gr{\"o}ller (1995).

The contributions of this paper are:
\begin{itemize}
\item Formulating the superparaboloid in a similar parametric and implicit form as the superellipsoids in \citejacklicbook, including derivations of its normal vector;
\item Extending the close-to-uniform sampling ideas in Pilu and Fisher (1995) to the 3D case of superellipsoids and superparaboloids;
\item Including both a tapering and bending transformations into the overall sampling framework
\end{itemize}

Taken together these contributions form a complete method for uniform sampling of superparaboloids and superellipsoids with tapering and bending deformations. We provide pseudocode for uniform sampling of superellipses/superellipsoids and superparabolas/superparaboloids in the Appendix.

\section{Superquadrics}

\section{Superquadrics}

\subsection{Superellipsoids}

The parametric formulation of the superellipsoids is directly taken from \citejacklicbook and we introduce our own formulation for the superparaboloid (Sec.~\ref{sec:superparaboloids_formulation}).

The superellipsoid parametric surface vector can be obtained from the spherical product of two superellipses

\begin{gather}
\label{eq:sq_superellipses}
\begin{split}
\bf{r}(\eta , \omega) = 
\begin{bmatrix}
 \cos^{\epsilon_1} \eta \\
 a_3  sin^{\epsilon_1} \eta
\end{bmatrix} \otimes 
\begin{bmatrix}
 a_1 \cos^{\epsilon_2} \omega \\
 a_2 \sin^{\epsilon_2} \omega
\end{bmatrix} \\
\begin{bmatrix}
 a_1 \cos^{\epsilon_1} \eta \cos^{\epsilon_2} \omega \\
 a_2 \cos^{\epsilon_1} \eta \sin^{\epsilon_2} \omega \\
 a_3 \sin^{\epsilon_1} \eta
\end{bmatrix}, \\
-\frac{\pi}{2} \leq \eta \leq \frac{\pi}{2} \\
-\pi \leq \omega < \pi
\end{split}
\end{gather}

with the implicit equation being

\begin{align*}
	\bigg( \bigg( \frac{x}{a_1}\bigg) ^\frac{2}{\epsilon_2} + \bigg( \frac{y}{a_2}\bigg) ^\frac{2}{\epsilon_2} \bigg) ^\frac{\epsilon_2}{\epsilon_1} + \bigg( \frac{z}{a_3}\bigg) ^\frac{2}{\epsilon_1} = 1
\end{align*}

where parameters $a_1$, $a_2$ and $a_3$ define the size of the superellipsoid in the $x$, $y$ and $z$ dimensions respectively; and  $\epsilon_1$ and $\epsilon_2$ control the shape (Fig.~\ref{fig:intro_superquadrics}). Note that by setting $a_1 = a_2 = a_3 = 1$ and $\epsilon_1 = \epsilon_2 = 1$ we get the unit sphere.

We can then build a function.

\begin{gather}
\label{eq:superellipsoid_func}
	F(\mathbf{x};\Lambda) = \bigg( \bigg( \frac{x}{a_1}\bigg) ^\frac{2}{\epsilon_2} + \bigg( \frac{y}{a_2}\bigg) ^\frac{2}{\epsilon_2} \bigg) ^\frac{\epsilon_2}{\epsilon_1} + \bigg( \frac{z}{a_3}\bigg) ^\frac{2}{\epsilon_1}
\end{gather}

where $\mathbf{x}$ is the vector $\mathbf{x} = (x,y,z)$ and $\Lambda = (a_1, a_2,a_3,\epsilon_1, \epsilon_2)$ is the parameter vector. The function above is called the \textit{inside-outside} function because it provides a way to tell if a point $\mathbf{x}$ is inside ($F<1$), on the surface ($F=1$) or outside ($F>1$) the superellipsoid \cite{JacklicSQBook}.

It is possible to extend $\Lambda$ to define a superellipsoid in general position and orientation in space. We use $3$ extra parameters $(px,py,pz)$ for the position of its central point, and $3$ more $(\theta, \phi, \psi)$  for the $ZYZ$ Euler angles that fully define its orientation.
Now we have

\begin{gather}
\label{eq:lambda_params}
\Lambda = (a1, a2,a3,\epsilon_1,\epsilon_2,\theta,\phi,\psi,px,py,pz)
\end{gather}

and are able to define a superellipsoid in general position and orientation with $11$ parameters.

\subsection{Superparaboloids}
\label{sec:superparaboloids_formulation}

We start by defining a superparabola in the following parametric form

\begin{align*}
\textbf{x}(u) =
\begin{bmatrix}
 u \\
 a_3 (u^\frac{2}{\epsilon_1} - 1)
\end{bmatrix}
\end{align*}

Then, analogously to Eq.~\ref{eq:sq_superellipses}, a superparaboloid is the spherical product of a superparabola with a superellipse

\begin{gather}
\label{eq:sq_superparabolas}
\begin{split}
\textbf{r}(u, \omega) =
\begin{bmatrix}
 u \\
 a_3 (u^\frac{2}{\epsilon_1} - 1)
\end{bmatrix} \otimes 
\begin{bmatrix}
 a_1 \cos^{\epsilon_2} \omega \\
 a_2 \sin^{\epsilon_2} \omega
\end{bmatrix} \\
\begin{bmatrix}
 a_1 u \cos^{\epsilon_2}\omega \\
 a_2 u \sin^{\epsilon_2}\omega \\
 a_3 (u^\frac{2}{\epsilon_1} - 1)
\end{bmatrix}, \\
0 \leq u \leq 1 \\
-\pi \leq \omega < \pi
\end{split}
\end{gather}

By solving for the surface vectors X, Y and Z we get the inside-outside function in a similar form to Eq.~\ref{eq:superellipsoid_func}

\begin{align*}
F(\mathbf{x};\Lambda) = \bigg( \bigg( \frac{x}{a_1}\bigg) ^\frac{2}{\epsilon_2} + \bigg( \frac{y}{a_2}\bigg) ^\frac{2}{\epsilon_2} \bigg) ^\frac{\epsilon_2}{\epsilon_1} - \bigg( \frac{z}{a_3}\bigg)
\end{align*}

with the lambda parameter vector (Eq.~\ref{eq:lambda_params}) defining analogous values for scale, shape, orientation and position.

\subsection{Deformations}

We also include two known extension to superquadrics: \textit{tapering} and \textit{bending}. We use the tapering deformation introduced in \citejacklicbook that linearly thins or expands the superquadric along its $z$ axis. Tapering requires two extra parameters $K_x$ and $K_y$ for tapering in the $x$ and $y$ directions (Sec.~\ref{sec:def_tapering}). Regarding bending we define our own deformation that requires one parameter $k$ for the curvature (Sec.~\ref{sec:def_bending}).

We then have three extra parameters, with a final lambda

\begin{align*}
\Lambda = (a1,a2,a3,\epsilon_1,\epsilon_2,\theta,\phi,\psi,K_x,K_y,k,px,py,pz)
\end{align*}

This is our final set of $14$ parameters to define a superquadric tapered or bent, and in general position and orientation. We combine our transformations (translation, rotation, bending and tapering) in the same order as in \citejacklicbook:

\begin{align*}
\mathit{Trans}(\mathit{Rot}(\mathit{Bend}(\mathit{Taper}(\bf{x}))))
\end{align*}

A deformation is defined as a function $D$ that directly modifies the global coordinates of the surface points

\begin{align*}
\bf{X}=\bf{D(X)}= 
\begin{bmatrix}
 X(x,y,z) \\
 Y(x,y,z) \\
 Z(x,y,z)
\end{bmatrix}
\end{align*}

\subsubsection{Tapering}
\label{sec:def_tapering}

We consider the tapering deformation as in \citejacklicbook, which allows us to taper a superquadric along the $z$ axis differently in $x$ and $y$ dimensions. We have $f_x(z)$ and $f_y(z)$ as the tapering functions along the respective axes. The tapering deformation is then a function of $z$ and we have the new surface vectors

\begin{gather}
\label{eq:taper_surf_vec}
\begin{aligned}
& X = f_x(z)x \\
& Y = f_y(z)y \\
& Z = z
\end{aligned}
\end{gather}

The two tapering functions are 

\begin{gather}
\label{eq:taper_funcs}
\begin{aligned}
& f_x(z) = \frac{K_x}{a_3} z + 1 \\
& f_y(z) = \frac{K_y}{a_3} z + 1 \\
\end{aligned}
\end{gather}

The parameters $K_x$ and $K_y$ control the amount and direction of tapering along each dimension and define them in the interval $-1 <= K_x,K_y <= 1$. For no tapering we set $Kx=Ky=0$. 

\subsubsection{Bending}
\label{sec:def_bending}

We create our own, simpler bending deformation that uses the circle function to deform the superquadric, which is bent positively on $X$ along $Z$. There is only one parameter defining the circle's radius

\begin{align*}
k \geq a_3
\end{align*}

Bending gives us the new surface vector components

\begin{gather}
\label{eq:bend_surf_vec}
\begin{aligned}
& X = x + (k - \sqrt{k^2 + z^2}) \\
& Y = y \\
& Z = z
\end{aligned}
\end{gather}

The maximum bending is when $k = a_3$ and we have no bending for $k \gg a_3$.

\section{Uniform Sampling}

\subsection{Point Sampling}

\subsubsection{Superellipsoids}
\label{sec:sample_superellipsoids}

For the uniform sampling of superellipsoids we extend the equations in Pilu and Fisher (1995) to the 3D case. Regarding transformations, in practice we did not need to derive equations taking them into account and instead we found a simple deformation made on the point cloud after sampling to be sufficient. That is, we sample the angles as if there was no deformation; create the point cloud from the sampled angles; and only then apply the tapering transformation to the point cloud. Although the tapering and bending are not isometries (i.e. do not preserve distance), this simpler method serves our practical purposes.

Pilu and Fisher (1995) derive an algorithm for sampling angles $\theta$ of a parametric superellipse,

\begin{align*}
\textbf{x}(\theta) = 
\begin{bmatrix}
 a \cos^{\epsilon}(\theta) \\
 b  \sin^{\epsilon}(\theta)
\end{bmatrix}
\end{align*}

so as to maintain a constant arc length between the points. They approximate the arclength between two points as a straight line connecting them

\begin{align*}
\textbf{D}^2(\theta) = |\textbf{x}(\theta+\Delta_{\theta}(\theta)) - \textbf{x}(\theta)|^2
\end{align*}

and approximate the right-hand side to first order

\begin{align*}
\textbf{D}^2(\theta) = \Big( \frac{\partial}{\partial \theta} (a \cos^\epsilon(\theta)) \Delta_{\theta}(\theta) \Big)^2 + \Big( \frac{\partial}{\partial \theta} (b \sin^\epsilon(\theta)) \Delta_{\theta}(\theta) \Big)^2
\end{align*}

then solve it for $\Delta_{\theta}(\theta)$ yielding

\begin{gather}
\label{eq:arclength_approx_superellipse}
\Delta_{\theta}(\theta) = \frac{\textbf{D}(\theta)}{\epsilon}\sqrt{\frac{\cos^2(\theta) \sin^2(\theta)}{a^2 \cos^{2\epsilon}(\theta) \sin^4(\theta) + b^2 \sin^{2\epsilon}(\theta) \cos^4(\theta)  }}
\end{gather}

The arclength $\textbf{D}(\theta)$ can be set to a constant and the $\theta$ angles are obtained by iteratively updating $\theta_i$ in a dual manner

\begin{align*}
\theta_i = \theta_{i-1} + \Delta_{\theta}(\theta_i), \hspace{0.5cm} \theta_0 = 0, \hspace{0.5cm} \theta_i < \frac{\pi}{2} \\
\theta_i = \theta_{i-1} - \Delta_{\theta}(\theta_i), \hspace{0.5cm} \theta_0 = \frac{\pi}{2}, \hspace{0.5cm} \theta_i > 0 \\
\end{align*}

The first incrementing up from $\theta=0$ while $\theta<\frac{\pi}{2}$ and the second incrementing down from $\theta=\frac{\pi}{2}$ while $\theta > 0$. The authors also derive a second equation in order to avoid singularities when $\theta$ is very close to $0$ or $\frac{\pi}{2}$:

\begin{align*}
\Delta_{\theta}(\theta)_{\theta \to 0} = \Big( \frac{\textbf{D}(\theta)}{b} - \theta^\epsilon \Big)^{\frac{1}{\epsilon}} - \theta \\
\Delta_{\theta}(\theta)_{\theta \to \frac{\pi}{2}} = \Big( \frac{\textbf{D}(\theta)}{a} - (\frac{\pi}{2} - \theta)^\epsilon \Big)^{\frac{1}{\epsilon}} - (\frac{\pi}{2} - \theta )
\end{align*}

Using these ideas from Pilu and Fisher (1995) all we need to do is adapt them to the 3D case, i.e., to both superellipses used for the spherical product of a superquadric (Eq.~\ref{eq:sq_superellipses}). For sampling the $\eta$ angles for the first superellipse we substitute

\begin{align*}
\theta = \eta \hspace{0.3cm} \epsilon=\epsilon_1 \hspace{0.3cm} a=a_1 \hspace{0.3cm} b=a_2
\end{align*}

and for the $\omega$ angles

\begin{align*}
\theta = \omega \hspace{0.3cm} \epsilon=\epsilon_2 \hspace{0.3cm} a=1 \hspace{0.3cm} b=a_3
\end{align*}

Since superellipsoids are symmetrical with respect to the three axis, we need only sample from $0$ to $\frac{\pi}{2}$ and then mirror the results.

\subsubsection{Superparaboloids}

In order to uniformly sample for a superparaboloid we sample for its superparabola and superellipsoid. For the superellipsoids we sample the same as in Sec.~\ref{sec:sample_superellipsoids}. For the superparabola we apply the same approximation (as in Sec.~\ref{sec:sample_superellipsoids}). We start with the superparabola parametric equation

\begin{align*}
\textbf{x}(u) = 
\begin{bmatrix}
 u \\
 a_3 (u^\frac{2}{\epsilon_1} - 1)
\end{bmatrix}
\end{align*}

and the arclength approximation

\begin{align*}
\textbf{D}^2(u) = |\textbf{x}(u+\Delta_{u}(u)) - \textbf{x}(u)|^2
\end{align*}

Approximating the right-hand side to first order and solving for $\Delta_{u}(u)$ yields

\begin{gather}
\label{eq:arclength_approx_superparabola}
\Delta_{u}(u) = \frac{\textbf{D}(u)}{\sqrt{\frac{4a_3^2}{\epsilon_1^2}u^{\frac{4}{\epsilon_1}-2} + 1 }}
\end{gather}

and the update incrementing from $0$ while $u \leq 1$.

\begin{align*}
u_i = u_{i-1} + \Delta_{u}(u_i), \hspace{0.5cm} u_0 = 0, \hspace{0.5cm} u_i \leq 1 \\
\end{align*}

Since the superparabola is symmetrical with respect to the $Y$ axis we need only sample from $u_i = 0$ to $1$ and then duplicate the points, changing the sign for the $X$ values. In order to sample the 3D superparaboloid we sample both the $u_i$ for the superparabola and the $\theta_i$ for a superellipse.

\subsection{Normal Sampling}

\subsubsection{Normal for Non-Deformed Surfaces}

We also obtain the normals at each sampled point. For superellipsoids without deformations we use the parametric normal vector derived in \citejacklicbook. The vector for the direction of the normals in terms of the components of the surface vector ($x$, $y$ and $z$) is given by

\begin{align*}
\textbf{n}(\eta , \omega) = 
\begin{bmatrix}
 \frac{1}{x} \cos^2 \eta \cos^2 \omega \\
 \frac{1}{y} \cos^2 \eta sin^2 \omega \\
 \frac{1}{z} \sin^2 \eta
\end{bmatrix}
\end{align*}

For superparaboloids without deformations we derive the normal vector below. We start with the tangent vectors along the coordinates' curves

\begin{align*}
\textbf{r}_{u}(u,\omega) = 
\begin{bmatrix}
 a_1 \cos^{\epsilon_2} \omega \\
 a_2 \sin^{\epsilon_2} \omega \\
 \frac{2 a_3}{\epsilon_1} u^{\frac{2}{\epsilon_1}-1}
\end{bmatrix} \\
\textbf{r}_{\omega}(u,\omega) = 
\begin{bmatrix}
 -a_1 u \epsilon_2 \sin\omega \cos^{\epsilon_2-1}\omega \\
 a_2 u \epsilon_2 \sin^{\epsilon_2-1}\omega \cos\omega \\
 0
\end{bmatrix}
\end{align*}

The cross product of the tangent vectors is

\begin{align*}
\begin{split}
\textbf{r}_{u}(u,\omega) \times \textbf{r}_{\omega}(u,\omega) = \\
\begin{bmatrix}
 -\frac{2 a_3 \epsilon_2}{\epsilon_1} a_1 u^{\frac{2}{\epsilon_1}} \sin^{\epsilon_2-1} \omega \cos \omega \\
 -\frac{2 a_3 \epsilon_2}{\epsilon_1} a_2 u^{\frac{2}{\epsilon_1}} \sin \omega \cos^{\epsilon_2-1} \omega \\
 a_1 a_2 \epsilon_2 u \sin^{\epsilon_2-1}\omega \cos^{\epsilon_2-1}\omega\\
 \end{bmatrix}
\end{split}
\end{align*}

If we define a scalar function

\begin{align*}
f(u,\omega) = -2 a_1 a_2 a_3 \frac{\epsilon_2}{\epsilon_1} u^{\frac{2}{\epsilon_1}-1}  \sin^{\epsilon_2-1}\omega \cos^{\epsilon_2-1}\omega
\end{align*}

we have the cross product as

\begin{align*}
\textbf{r}_{u}(u,\omega) \times \textbf{r}_{\omega}(u,\omega) =
f(u,\omega) 
\begin{bmatrix}
 \frac{1}{a_1} u \cos^{2-\epsilon_2} \omega \\
 \frac{1}{a_2} u \sin^{2-\epsilon_2} \omega \\ 
 -\frac{1}{a_3} \frac{\epsilon_1}{2} u^{2-\frac{2}{\epsilon_1}}
 \end{bmatrix}
\end{align*}

With this we get the dual superparaboloid (similarly to the dual superquadric in \cite{Barr1981}). By dropping the scalar function, the normal vector of the original superparaboloid becomes the surface vector for the dual one

\begin{align*}
\textbf{n}_d(u,\omega) =
\begin{bmatrix}
 \frac{1}{a_1} u \cos^{2-\epsilon_2} \omega \\
 \frac{1}{a_2} u \sin^{2-\epsilon_2} \omega \\ 
 -\frac{1}{a_3} \frac{\epsilon_1}{2} u^{2-\frac{2}{\epsilon_1}}
 \end{bmatrix}
\end{align*}

The normal vector can also be represented in terms of the components of the surface vector \cite{JacklicSQBook}

\begin{align*}
\textbf{n}_d(u,\omega) =
\begin{bmatrix}
 \frac{1}{x} u \cos^{2} \omega \\
 \frac{1}{y} u \sin^{2} \omega \\ 
 -\frac{1}{z} \frac{\epsilon_1}{2} u^{\frac{2}{\epsilon_1}}
 \end{bmatrix}
\end{align*}

\subsubsection{Normal for Deformed Surfaces}

It is possible to obtain, for the deformed surface, the normal vector $\bf{n_t}$ at each point from the original surface normal vector $\bf{n_o}$ by applying a transformation matrix $\bf{T}$ \cite{Barr1984,JacklicSQBook}.

\begin{align*}
\bf{n_t}(\eta , \omega) = T \bf{n_o}(\eta, \omega) \\
\bf{T} = \det \bf{J}  \bf{J}^{-1 \it{T}}
\end{align*}

The same can also be made for a superparaboloid normal vector $\bf{n_o}(u, \omega)$. As for the matrix $\bf{J}$, it is the Jacobian of $\bf{D}$, given by

\begin{gather}
\label{eq:jacob_deform}
\bf{J}(\bf{x}) = 
\begin{bmatrix}
	\frac{\partial \bf{X}}{\partial x} & \frac{\partial \bf{X}}{\partial y} & \frac{\partial \bf{X}}{\partial z} \\  
	\frac{\partial \bf{Y}}{\partial x} & \frac{\partial \bf{Y}}{\partial y} & \frac{\partial \bf{Y}}{\partial z} \\
	\frac{\partial \bf{Z}}{\partial x} & \frac{\partial \bf{Z}}{\partial y} & \frac{\partial \bf{Z}}{\partial z}
\end{bmatrix}
\end{gather}

Therefore we need only derive the Jacobian of a given transformation in order to get the normals. The tapering Jacobian is provided in \citejacklicbook. In the following sections we derive normal transformation matrices for tapering and bending.

\subsubsection{Tapering}

By substituting equations \ref{eq:taper_surf_vec} and \ref{eq:taper_funcs} into \ref{eq:jacob_deform}, and taking the partial derivatives, we get the Jacobian $\bf{J_t}$  for the tapering deformation as

\begin{align*}
\bf{J_t}(\bf{x}) = 
\begin{bmatrix}
	f_x(z) & 0 & \frac{\partial f_x(z)}{\partial z} x \\  
	0 & f_x(z) & \frac{\partial f_y(z)}{\partial z} y \\
	0 & 0 & 1
\end{bmatrix}
\end{align*}

we then have

\begin{align*}
\det \bf{J_t}(\bf{x}) = f_xf_y \\
\bf{J_t}(\bf{x})^{-1 \it{T}} = 
\begin{bmatrix}
	\frac{1}{f_x} & 0 & 0 \\  
	0 & \frac{1}{f_y} & 0 \\
	-\frac{f\prime _x}{f_x} x & -\frac{f\prime _y}{f_y} y & 1
\end{bmatrix}
\end{align*}

The normal tapering transformation $\bf{T}$ is then

\begin{align*}
\bf{T} = 
\begin{bmatrix}
	f_y & 0 & 0 \\  
	0 & f_x & 0 \\
	-\frac{f\prime _x}{f_y} x & -\frac{f\prime _y}{f_x} y & f_xf_y
\end{bmatrix}
\end{align*}

It is interesting to note that by considering tapering parameters $K_x = 0$ and $K_y = 0$, $f_x$ and $f_y$ become $1$ and ${f^{\prime}_x}$ and ${f^{\prime}_y}$ become $0$. Thus the transformation $\mathbf{T}$ becomes the identity matrix, keeping the original normal vector unchanged.

\subsubsection{Bending}
\label{sec:bending}

The bending Jacobian is very simple and given by substituting equation \ref{eq:bend_surf_vec} into \ref{eq:jacob_deform}

\begin{align*}
\bf{J_b} = 
\begin{bmatrix}
	1 & 0 & -\frac{z}{\sqrt{k^2+z^2}} \\  
	0 & 1 & 0 \\ 
	0 & 0 & 1 \\  
\end{bmatrix}
\end{align*}

and we get out normal transformation matrix as

\begin{align*}
\bf{T} = 
\begin{bmatrix}
	1 & 0 & 0 \\  
	0 & 1 & 0 \\
	\frac{z}{\sqrt{k^2+z^2}} & 0 & 1
\end{bmatrix}
\end{align*}

then we note that the transformation converges to identity as $k$ increases since 

\begin{align*}
\lim_{k\to\infty} \frac{z}{\sqrt{k^2+z^2}} = 0
\end{align*}

\subsubsection{Tapering Singularities}

Since we only care for the direction of the normal vectors we can drop the determinant multiplication. We then have our transformation matrix $\bf{T}= \bf{J_t}(\bf{x})^{-1 \it{T}}$ \cite{Barr1984}.

\begin{align*}
\bf{T} = 
\begin{bmatrix}
	\frac{1}{f_x} & 0 & 0 \\  
	0 & \frac{1}{f_y} & 0 \\
	-\frac{f\prime _x}{f_x} x & -\frac{f\prime _y}{f_y} y & 1
\end{bmatrix}
\end{align*}

For tapering, $\bf{T}$ has positive and negative infinities whenever $f_x(z) = 0$ or $f_y(z) = 0$. To avoid this when implementing, we can update $f_x(z)$ and $f_y(z)$ before calculating $\bf{T}$.

\begin{align*}
f_{x/y}(z) = 
\left\{
	\begin{array}{ll}
		\epsilon  & \mbox{if } f_{x/y}(z) = 0 \\
		f_{x/y}(z) & \mbox{if } f_{x/y}(z) \neq 0
	\end{array}
\right.
\end{align*}

Where $\epsilon$ can be defined to be a very small number. After transforming the original normal vector we can always obtain the unit normal vector by dividing the output vector by its magnitude.

\subsection{Results}

All experiments and figures in this paper were generated in the same desktop computer: Intel(R) Core(TM) i5-3470 CPU @ 3.20GHz. 

\subsubsection{Quantitative}
\label{sec:res_quant}

\begin{figure}[t!] 
  \begin{subfigure}[b]{0.5\linewidth}
    \centering
    \includegraphics[width=0.7\linewidth]{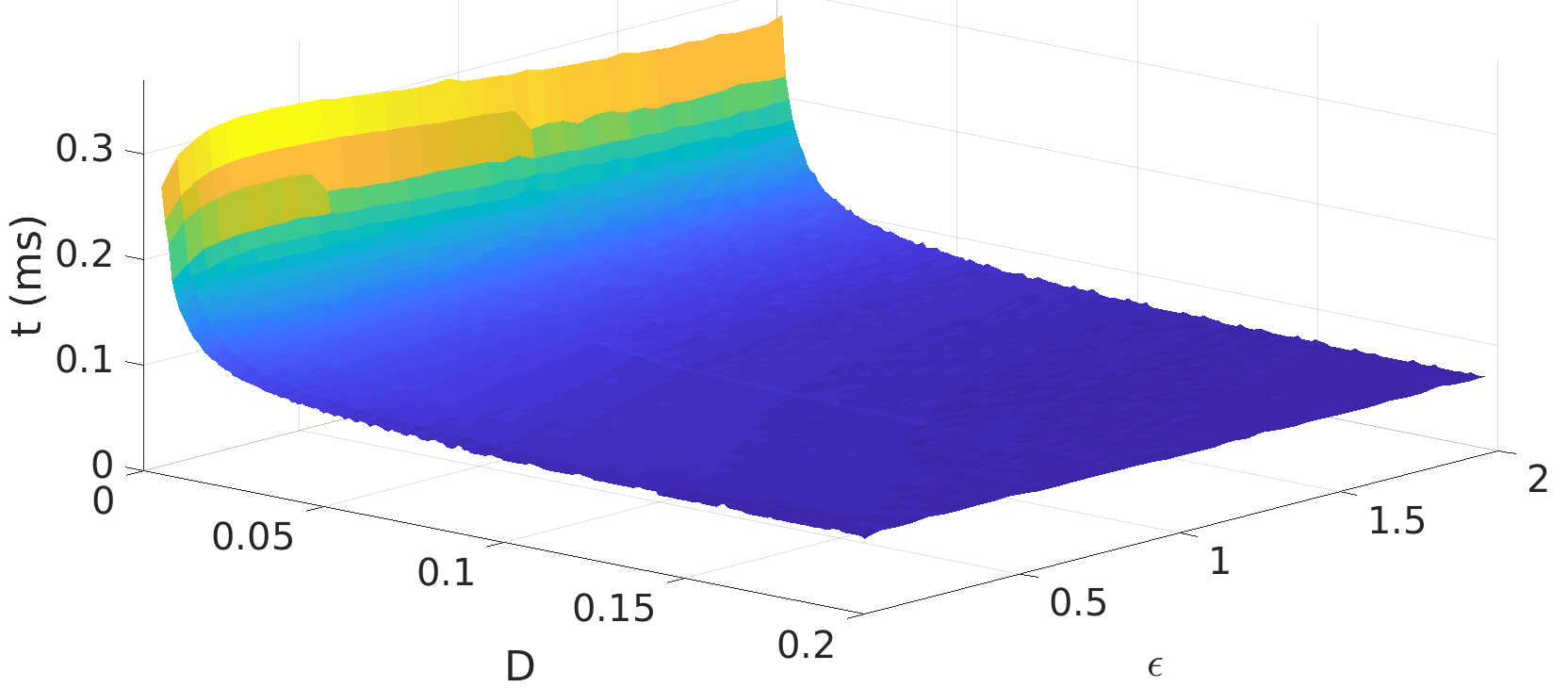} 
    \caption{Superellipses. Sampled $1,616$ to $32$ points.} 
    \label{fig:sampling_superellipses} 
    \vspace{2ex}
  \end{subfigure}
  \begin{subfigure}[b]{0.5\linewidth}
    \centering
    \includegraphics[width=0.6\linewidth]{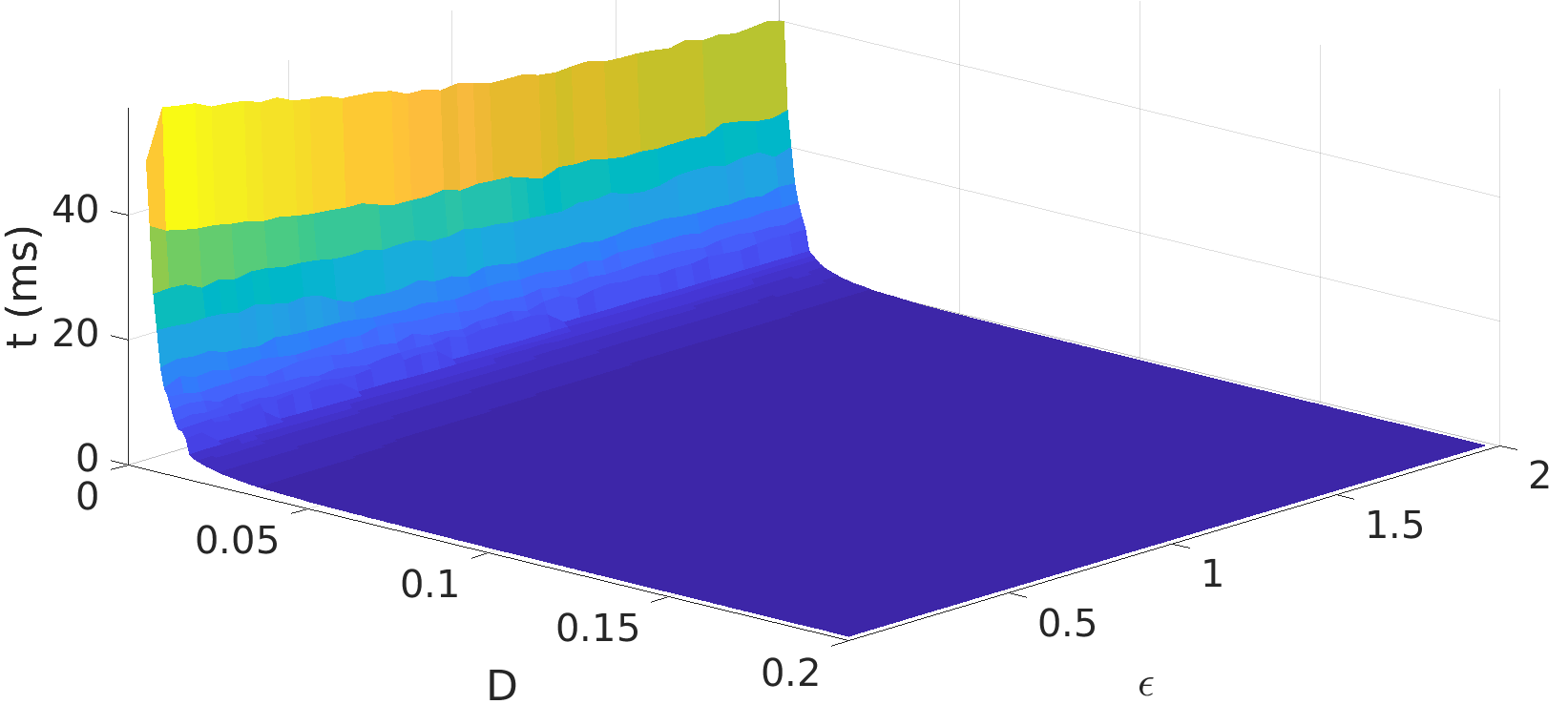} 
    \caption{Superellipsoids. Sampled $1,021,312$ to $512$ points.} 
    \label{fig:sampling_superellipsoids} 
    \vspace{2ex}
  \end{subfigure} 
  \begin{subfigure}[b]{0.5\linewidth}
    \centering
    \includegraphics[width=0.7\linewidth]{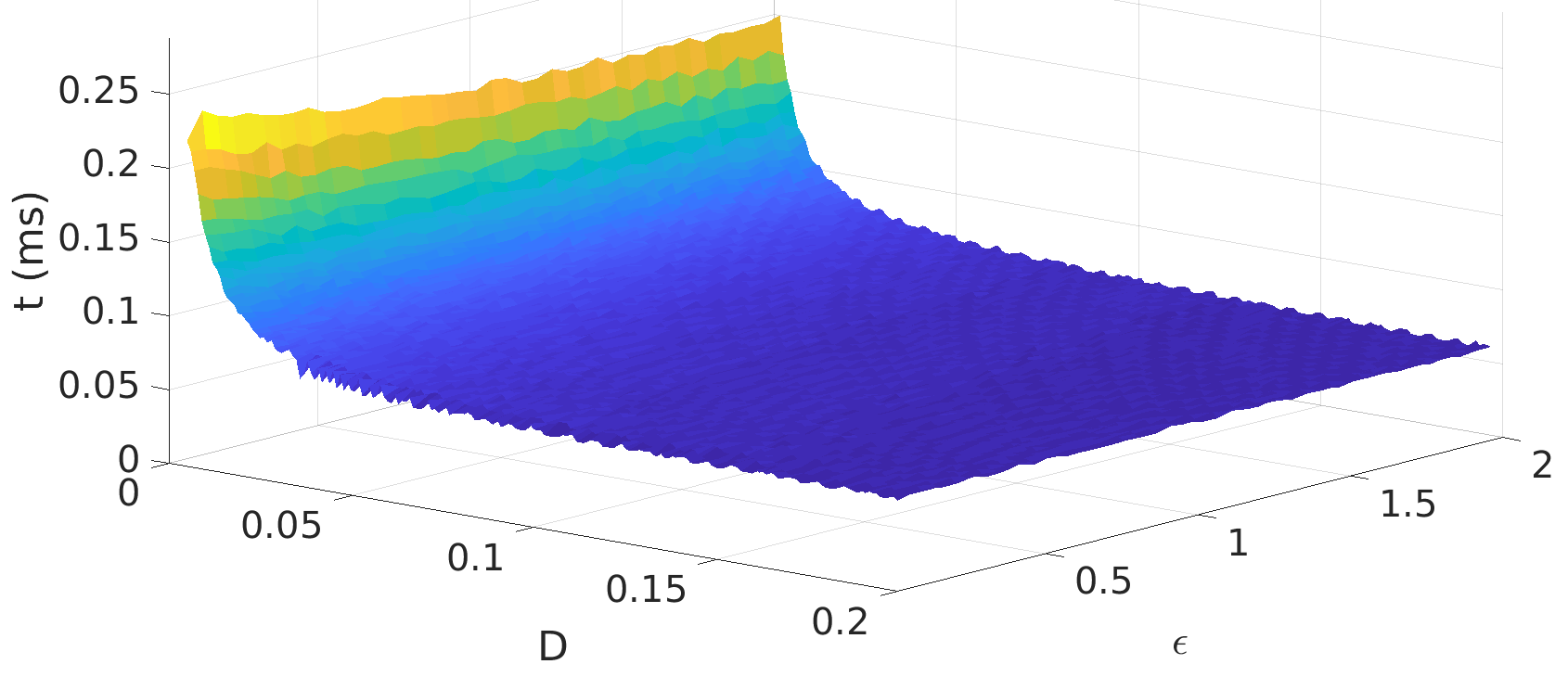} 
    \caption{Superparabolas. Sampled $3,681$ to $11$ points.} 
    \label{fig:sampling_superparabolas}
  \end{subfigure}
  \begin{subfigure}[b]{0.6\linewidth}
    \centering
    \includegraphics[width=0.6\linewidth]{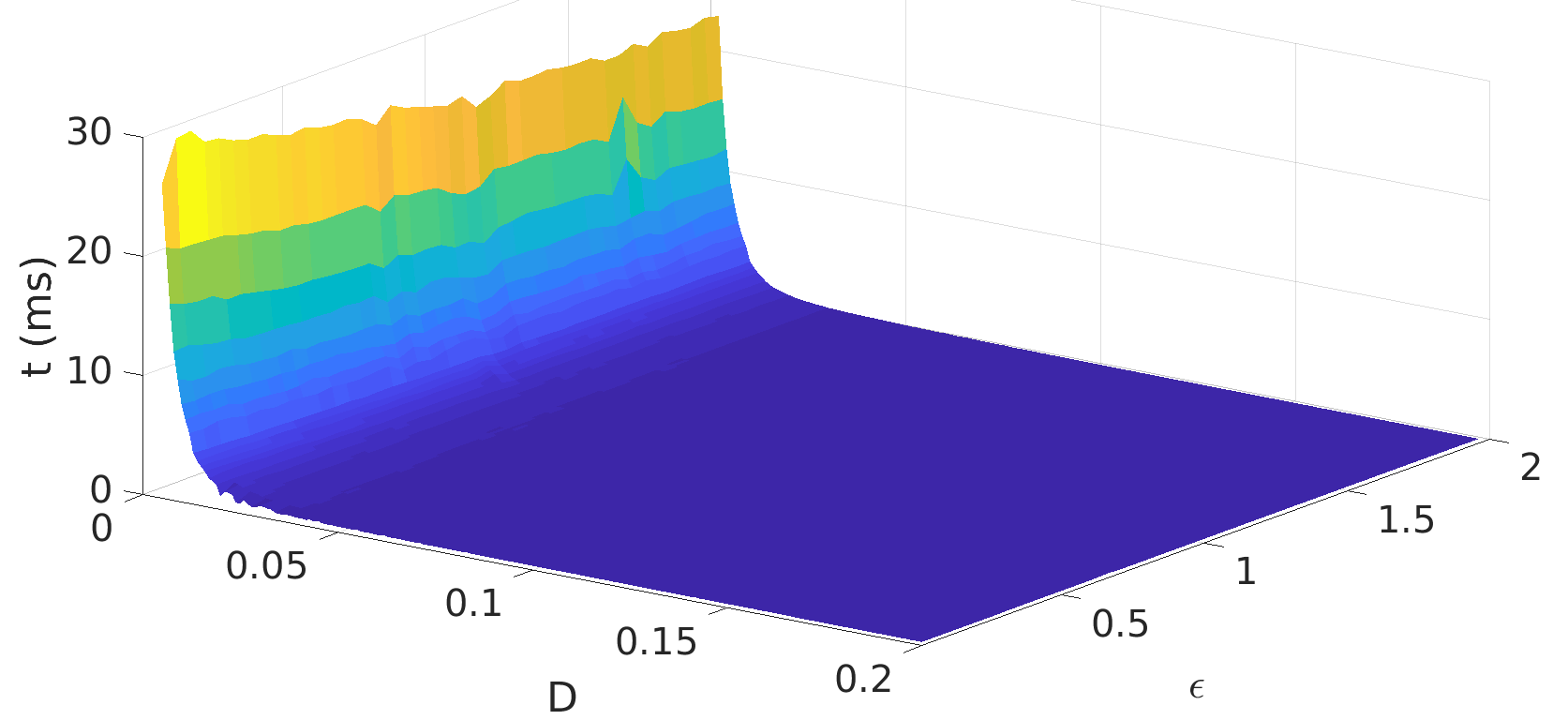} 
    \caption{Superparaboloids. Sampled $498,016$ to $192$ points.} 
    \label{fig:sampling_superparaboloids}
  \end{subfigure} 
  \vspace{0.1cm}
  \caption{Sampling time (ms) when varying $\epsilon$ and $D$ parameters.}
  \label{fig:sampling_times}
\end{figure}

For all quantitative experiments below, times are reported in milliseconds as the median over $1000$ trials and we vary $\epsilon$ from $0.1$ to $2$ in steps of $0.05$ and $D$ from $0.005$ to $0.2$ in steps of $0.001$. In Fig~\ref{fig:sampling_superellipses} we show the sampling times for sampling superellipses; in Fig~\ref{fig:sampling_superparabolas} for superparabolas; in Fig~\ref{fig:sampling_superellipsoids}, for superellipsoids; and in Fig~\ref{fig:sampling_superparaboloids}, for superparaboloids. 

\begin{figure}[h]
    \centering
    \includegraphics[width=0.5\linewidth]{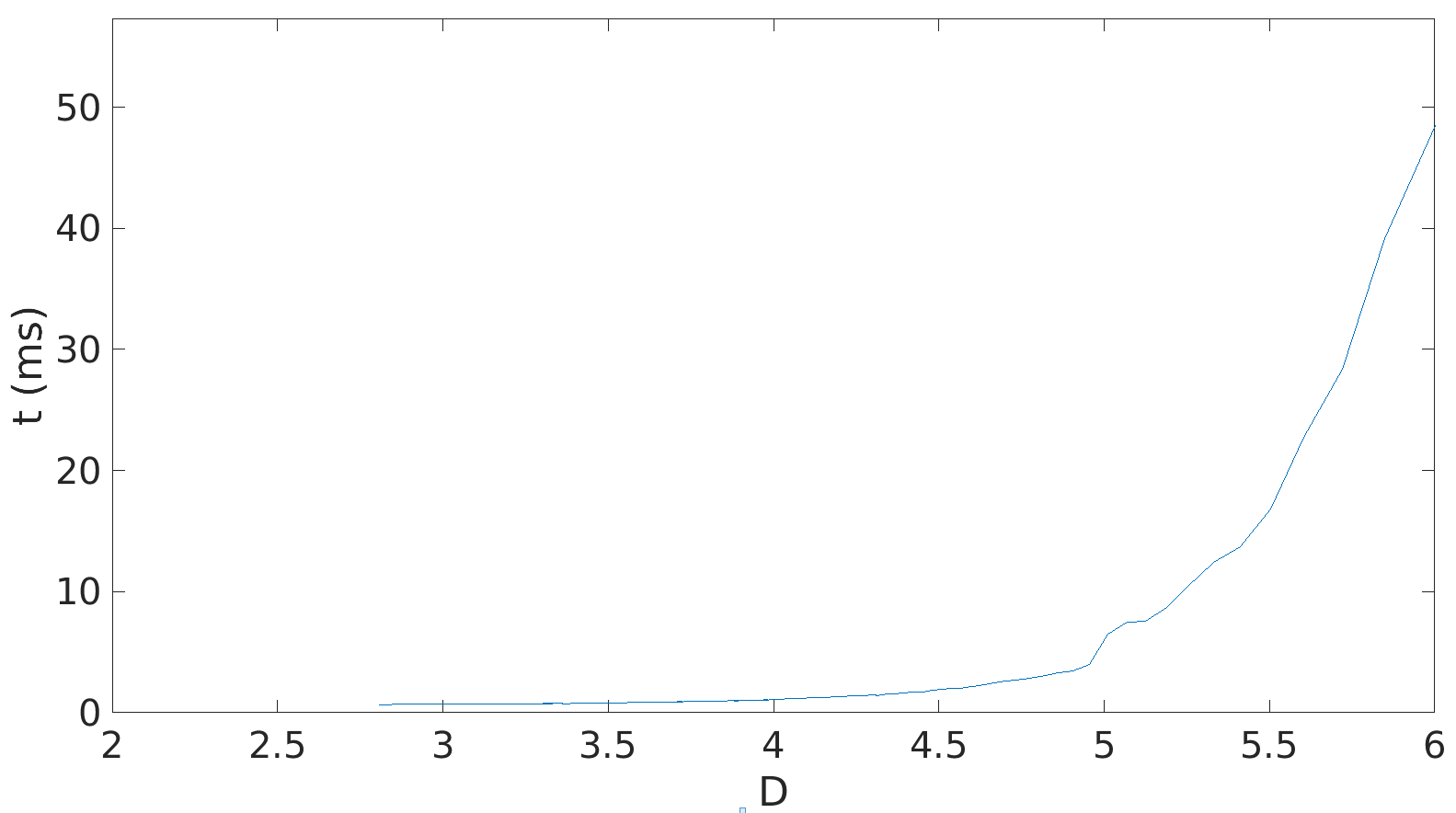} 
    \caption{Superellipsoids sampling time (ms) against the $log_{10}$ number of points, from $512$ to $1,021,312$.} 
    \label{fig:superellipsoids_npts} 
\end{figure}

All sampling results show an exponential decrease in time as $D$ gets larger and almost constant with respect to $\epsilon$. 
In Fig.~\ref{fig:superellipsoids_npts} we perform a cut on the results of Fig~\ref{fig:sampling_superellipsoids} and show how the sampling time varies with the $log_{10}$ number of points sampled for superellipsoids.

\subsubsection{Qualitative}

\begin{figure}[h!] 
  \begin{subfigure}[b]{0.5\linewidth}
    \centering
    \includegraphics[width=0.5\linewidth]{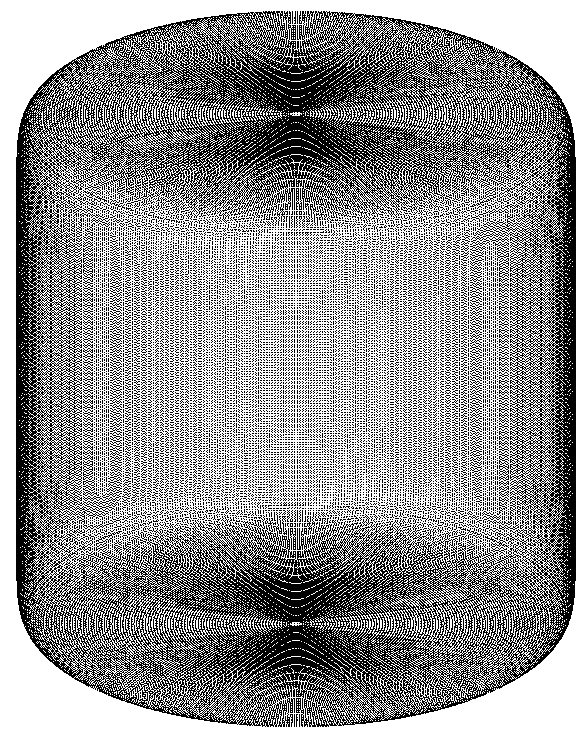} 
    \caption{$\epsilon_1=0.1$ $\epsilon_2=1$} 
    \label{fig:superellipsoid_01_1} 
    \vspace{2ex}
  \end{subfigure}
  \begin{subfigure}[b]{0.5\linewidth}
    \centering
    \includegraphics[width=0.5\linewidth]{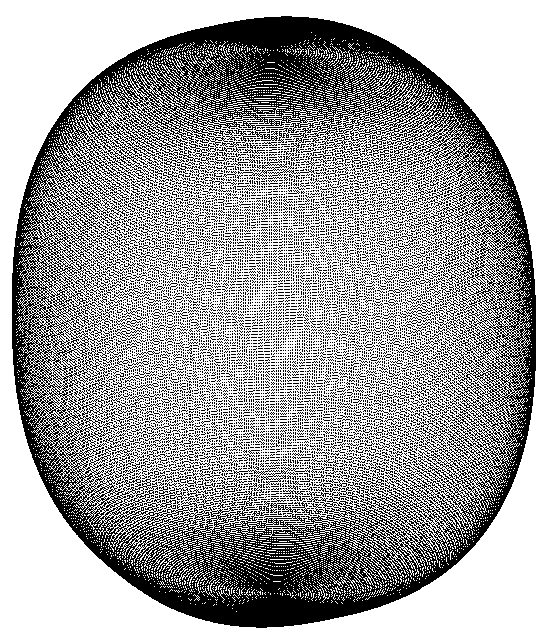} 
    \caption{$\epsilon_1=0.65$ $\epsilon_2=0.65$} 
    \label{fig:superegg} 
    \vspace{2ex}
  \end{subfigure} 
  \begin{subfigure}[b]{0.5\linewidth}
    \centering
    \includegraphics[width=0.5\linewidth]{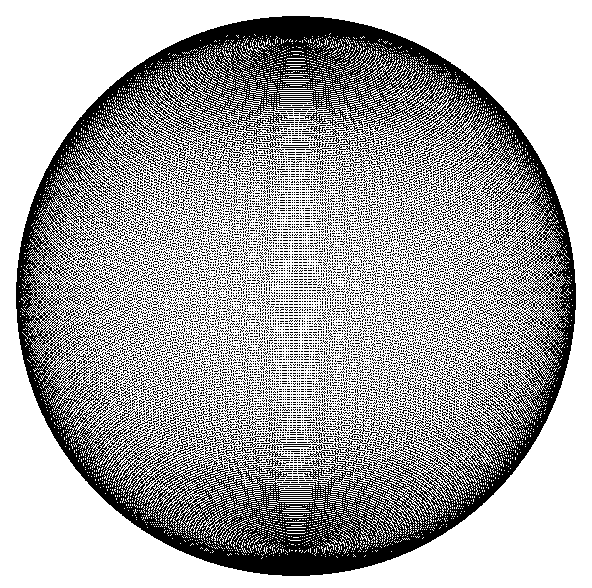} 
    \caption{$\epsilon_1=1$ $\epsilon_2=1$} 
    \label{fig:superellipsoid_1_1} 
  \end{subfigure}
  \begin{subfigure}[b]{0.5\linewidth}
    \centering
    \includegraphics[width=0.5\linewidth]{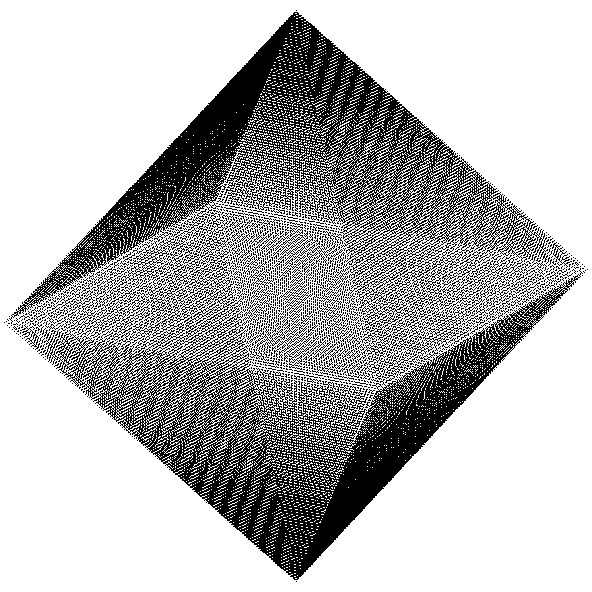} 
    \caption{$\epsilon_1=2$ $\epsilon_2=2$} 
    \label{fig:superellipsoid_2_2} 
  \end{subfigure} 
  \vspace{0.1cm}
  \caption{Different superellipsoids, including the superegg (b).}
  \label{fig:res_superellipsoids}
\end{figure}

\begin{figure}[h!] 
  \begin{subfigure}[b]{0.5\linewidth}
    \centering
    \includegraphics[width=0.5\linewidth]{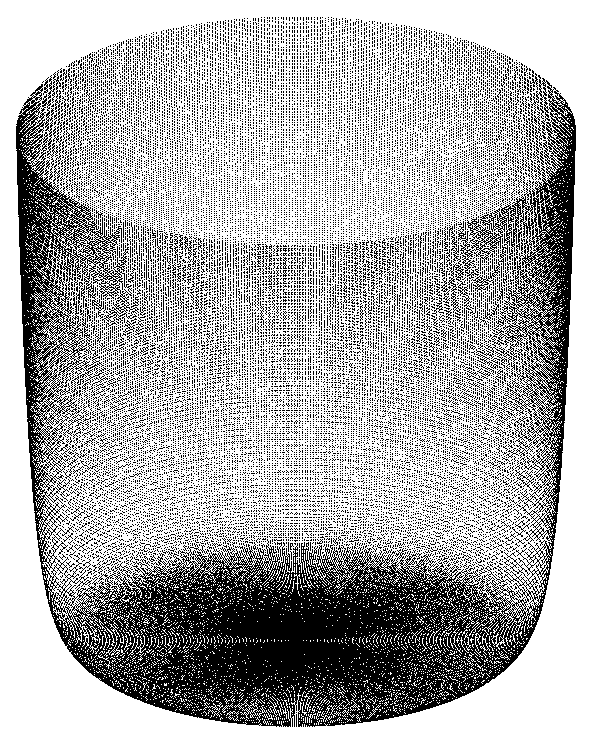} 
    \caption{$\epsilon_1=0.1$ $\epsilon_2=1$} 
    \label{fig:superparaboloid_01_1} 
    \vspace{2ex}
  \end{subfigure}
  \begin{subfigure}[b]{0.5\linewidth}
    \centering
    \includegraphics[width=0.5\linewidth]{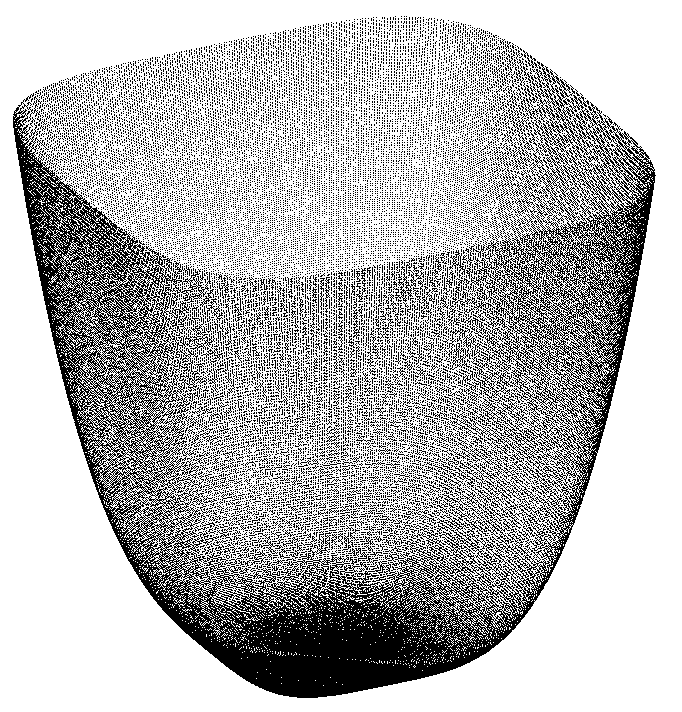} 
    \caption{$\epsilon_1=0.5$ $\epsilon_2=0.5$} 
    \label{fig:superparaboloid_05_05} 
    \vspace{2ex}
  \end{subfigure} 
  \begin{subfigure}[b]{0.5\linewidth}
    \centering
    \includegraphics[width=0.5\linewidth]{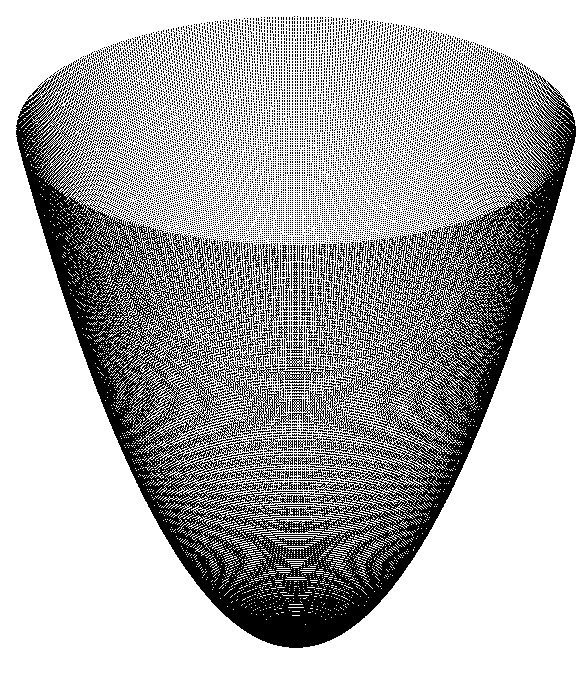} 
    \caption{$\epsilon_1=1$ $\epsilon_2=1$} 
    \label{fig:superparaboloid_1_1} 
  \end{subfigure}
  \begin{subfigure}[b]{0.5\linewidth}
    \centering
    \includegraphics[width=0.5\linewidth]{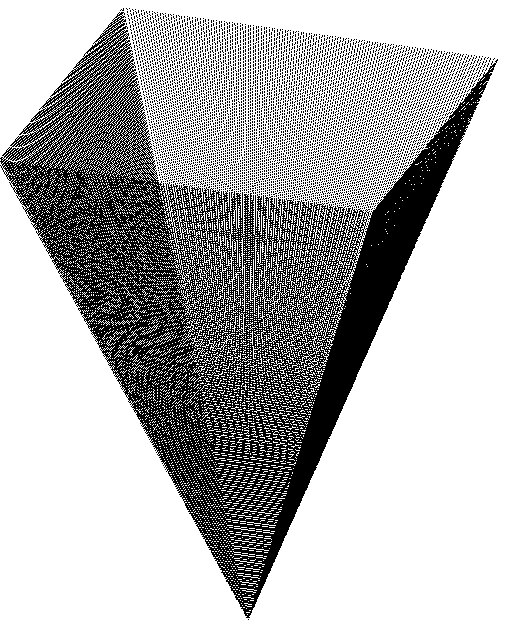} 
    \caption{$\epsilon_1=2$ $\epsilon_2=2$} 
    \label{fig:superparaboloid_2_2} 
  \end{subfigure} 
  \vspace{0.1cm}
  \caption{Different superparaboloids}
  \label{fig:res_superparaboloids}
\end{figure}

\begin{figure}[h!] 
  \begin{subfigure}[b]{0.5\linewidth}
    \centering
    \includegraphics[width=0.5\linewidth]{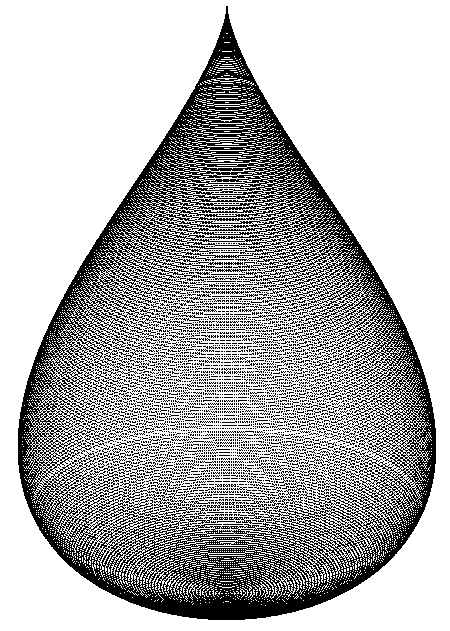} 
    \caption{Drop of water} 
    \label{fig:taper_1_1_1_1_1_m1_m1} 
    \vspace{2ex}
  \end{subfigure}
  \begin{subfigure}[b]{0.5\linewidth}
    \centering
    \includegraphics[width=0.4\linewidth]{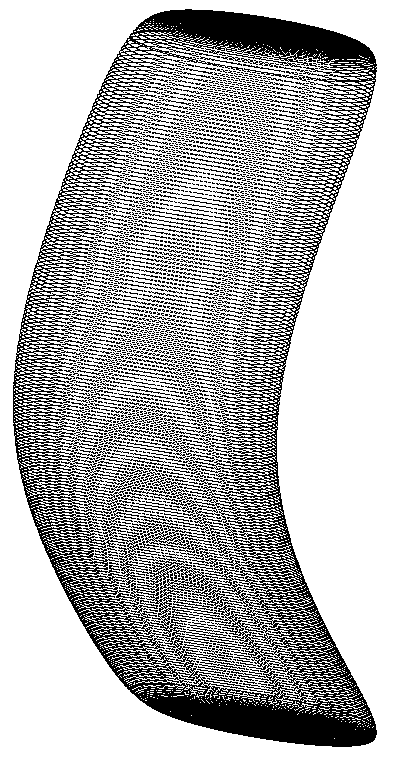} 
    \caption{Mug handle} 
    \label{fig:bent_1_3_7__02_1_4} 
    \vspace{2ex}
  \end{subfigure} 
  \begin{subfigure}[b]{0.5\linewidth}
    \centering
    \includegraphics[width=0.5\linewidth]{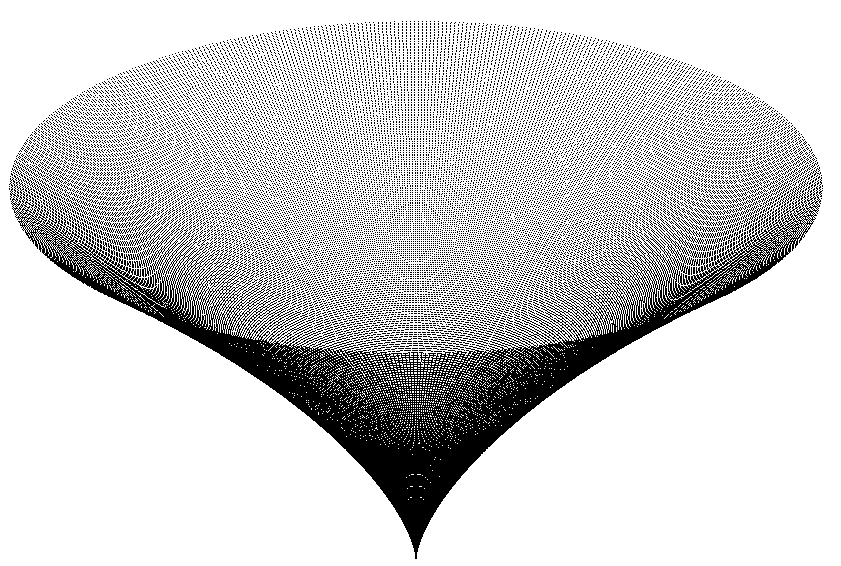} 
    \caption{Funnel} 
    \label{fig:sp_1_1_1_1_1_1_1} 
  \end{subfigure}
  \begin{subfigure}[b]{0.5\linewidth}
    \centering
    \includegraphics[width=0.5\linewidth]{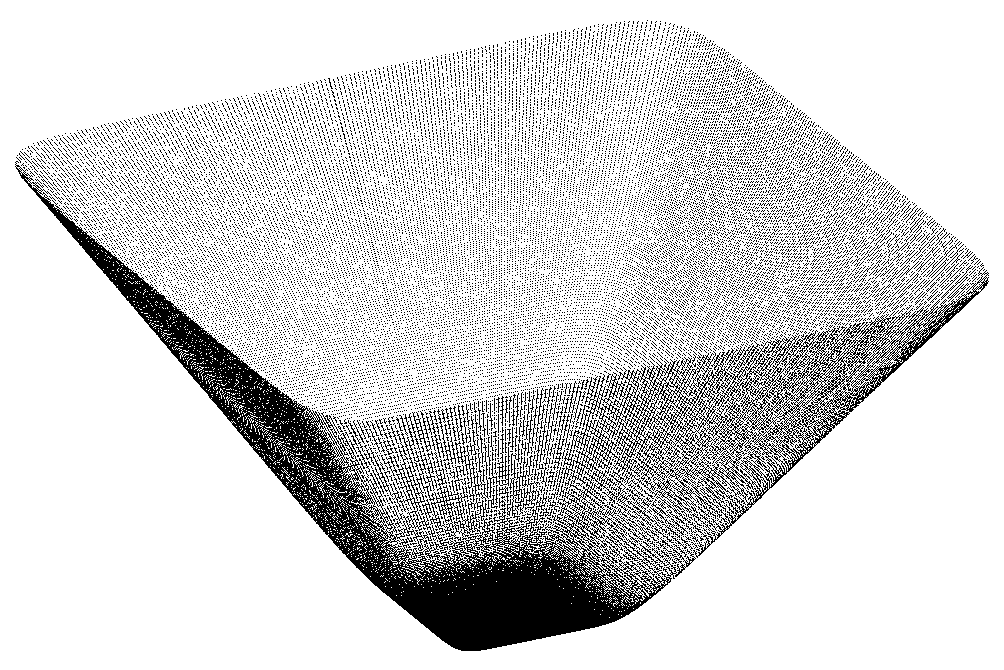} 
    \caption{Noodle bowl} 
    \label{fig:sp_12_12_1_02_02_05_05} 
  \end{subfigure} 
  \vspace{0.1cm}
  \caption{Examples of possible object or part modelling using different parameters with superellipsoids (a) and (b) and superparaboloids (c) and (d)}
  \label{fig:res_showcase}
\end{figure}

Here we show some qualitative results of our approach. In Fig.~\ref{fig:res_superellipsoids} we show sampled superellipsoid, for different values of $\epsilon_1$ and $\epsilon_2$, all with parameters $a_1=a_2=a_3=1$ and no bending or tapering. In Fig.~\ref{fig:res_superparaboloids} we show sampled superparaboloids, for different values of $\epsilon_1$ and $\epsilon_2$, all with parameters $a_1=a_2=a_3=1$ also with no bending or tapering. We also showcase a few examples of possible object or part modelling in Fig.~\ref{fig:res_showcase} as work in computer vision and robotics have used superquadrics as a compact representation of everyday objects
\cite{Andreopoulos2012,Krivic2004105,JacklicSQBook,BiegelbauerICRA2007,Page2003,Varadarajan2011b,Duncan2013,Drews2010,Strand2010,Guo2014,Cocias2012,Aleotti2012}.

We achieve close-to-uniform results for a great variety of shapes. Our implementation works well with $0 \leq \epsilon_1 \leq 2$, $0 \leq \epsilon_2 \leq 2$ and $\frac{max(a1,a2,a3)}{min(a1,a2,a3)} \leq 10$. Even within these limits it is possible to get many different shapes and sizes.

\section{Conclusion}

In this paper we have presented, extended, derived and implemented ideas for uniform sampling of superellipsoids and superparaboloids, including two deformations: tapering and bending. Our work builds heavily on \citejacklicbook and Pilu and Fisher (1995). We go beyond by introducing superparaboloids and 3D sampling in a complete framework with tapering and bending. In the future we plan to extend our work to include superhyperboloids (of one and two sheets) and supertoroids. This extension should not prove itself difficult if one follows the ideas in here. The work is limited in that the sampling could be improved further and there are still problems with sampling highly cubic superellipsoids with $\epsilon_1 < 0.1$ and solids with scale parameter proportion larger than $10$. This limitation can be seen in Fig.~\ref{fig:naive_vs_unif_cube} and in future work we plan to improve the model to allow sampling for very small $\epsilon$.

The sampling method is fast and the results are very close to uniform. We hope this paper may serve as a starting point for those interested in generating point clouds and normals from superquadrics. One interesting use of the uniform sampling is to provide a way of measuring the fitting quality of a superquadric to a point cloud \textit{[Removed for blind review]}. If one performs `recovery' of a superquadric from a point cloud \cite{JacklicSQBook} it is possible to perform an Euclidean distance between its points and a given point cloud to measure how well it represents the points; this is in contrast to using the inside-outside function as a measure.

\section*{ACKNOWLEDGEMENTS}
\noindent 
I would like to thank Frank Guerin for helping with text revision and ideas on how to structure the paper.

\bibliographystyle{IEEEtran}
\bibliography{library}

\clearpage

\section*{\uppercase{Appendix: Pseudocode}}
\label{app:pseudocode}
\noindent 

\begin{algorithm}
\caption{SuperParaboloidSampler}\label{alg:superparaboloid}
\begin{algorithmic}[1]
\Procedure{SuperParaboloid}{$a,b,c,\epsilon_1,\epsilon_2,D$}
\State $U \gets SampleSuperParabola(1,c,\epsilon_1,D)$ \Comment{List of all sampled $u$ parameters}
\State $\Omega \gets SampleSuperEllipse(a,b,\epsilon_2,D)$ \Comment{List of all sampled $\omega$ parameters}
\State $N_{\Omega} \gets length(\Omega)$ \Comment{Number of sampled $\omega$}
\State $X \gets a*U^T \cdot cos(\Omega)^{\epsilon_2}*$ \Comment{Get X component of surface vector}
\State $Y \gets a*U^T \cdot sin(\Omega)^{\epsilon_2}*$ \Comment{Get Y component of surface vector}
\State $Z \gets 2*c*\textbf{1}_{N_{\Omega}\times1}\cdot (U^2)^{\frac{1}{\epsilon_1}}$ \Comment{Get Z component of surface vector}
\State $X \gets concat(X,-X)$ \Comment{Series of concats, exploiting symmetry of superparaboloid...}
\State $X \gets concat(X,X)$ \Comment{...to generate final one by mirroring the sampled part}
\State $Y \gets concat(Y,Y)$
\State $Y \gets concat(Y,-Y)$
\State $Z \gets concat(Z,Z)$
\State $Z \gets concat(Z,Z)$
\State \Return (X,Y,Z)
\EndProcedure

\Procedure{SuperParabola}{$a,b,\epsilon,D$}
\State $U \gets SampleSP(a,b,\epsilon,D)$
\State $X \gets a*U$ \Comment{Get X component of surface vector}
\State $Y \gets b*(U^2)^{\frac{1}{\epsilon}}$ \Comment{Get Y component of surface vector}
\State $X \gets concat(X,-X)$ \Comment{Exploit symmetry to generate other half}
\State $Y \gets concat(Y,Y)$ \Comment{Exploit symmetry to generate other half}
\State \Return (X,Y)
\EndProcedure

\Procedure{SampleSP}{$a,b,\epsilon,D$}
\State $U(1) = 0$
\State $N = 1$
\While{$U < 1$}
\State $u_{next} \gets UpdateU(U(N),a,b,\epsilon,D)$
\State $N = N + 1$
\State $U(N) = u_{next}$
\EndWhile
\State \Return $\Theta$
\EndProcedure

\Procedure{UpdateU}{$u,a,b,\epsilon,D$}
\State $\Delta_{u}(u) =\gets \frac{D}{\sqrt{\frac{4a_3^2}{\epsilon_1^2}u^{\frac{4}{\epsilon_1}-2} + 1 }}$ \Comment{Approximate arclength  \refabeqp{arclength_approx_superparabola}}
\State \Return $u + \Delta_u(u)$
\EndProcedure
\end{algorithmic}
\end{algorithm}

\begin{algorithm}[h!]
\caption{SuperEllipsoidSampler}\label{alg:superellipsoid}
\begin{algorithmic}[1]
\Procedure{SuperEllipsoid}{$a,b,c,\epsilon_1,\epsilon_2,D$}
\State $H \gets SampleSuperEllipse(1,c,\epsilon_1,D)$ \Comment{List of all sampled $\eta$ parameters}
\State $\Omega \gets SampleSuperEllipse(a,b,\epsilon_2,D)$ \Comment{List of all sampled $\omega$ parameters}
\State $N_{\Omega} \gets length(\Omega)$ \Comment{Number of sampled $\omega$}
\State $X \gets [ ]$
\State $Y \gets [ ]$
\State $Z \gets [ ]$
\For{\texttt{<i=-1;i<=1;i+=2>}} \Comment{Tripled-nested for loops to get all 8 parts}
	\For{\texttt{<j=-1;j<=1;j+=2>}}
		\For{\texttt{<k=-1;k<=1;k+=2>}}
			\State $\cos_{\Omega} \gets \cos(j*\Omega)$
			\State $\sin_{\Omega} \gets \sin(j*\Omega)$
			\State $\cos_H \gets \cos(k*H)$
			\State $\sin_H \gets \sin(k*H)$
			\State $X_{next} \gets i*a* \cos^{\epsilon_2}_{\Omega}*\cos^{\epsilon_1}_H$ \Comment{Get X component of surface vector}
			\State $Y_{next} \gets i*b* \sin^{\epsilon_2}_{\Omega}*\cos^{\epsilon_1}_H$ \Comment{Get Y component of surface vector}
			\State $Z_{next} \gets i*c*\textbf{1}_{N_{\Omega}\times1}\cdot\sin_H$ \Comment{Get Z component of surface vector}
			\State $X \gets concat(X,X_{next}(:))$ \Comment{Get column vector of $X_{next}$ and add it to $X$}
			\State $Y \gets concat(Y,Y_{next}(:))$ 
			\State $Z \gets concat(Z,Z_{next}(:))$ 
		\EndFor
	\EndFor
\EndFor
\State \Return (X,Y,Z)
\EndProcedure
\end{algorithmic}
\end{algorithm}

\begin{algorithm}[h!]
\caption{SuperEllipseSampler}\label{alg:superellipse}
\begin{algorithmic}[1]
\Procedure{SuperEllipse}{$a,b,\epsilon,D$}
\State $\Theta \gets SampleSE(a,b,\epsilon,D)$
\State $X \gets a*cos^{\epsilon}(\Theta)$ \Comment{Get X component of surface vector}
\State $Y \gets b*sin^{\epsilon}(\Theta)$ \Comment{Get Y component of surface vector}
\State $X \gets concat(X,-X)$ \Comment{Exploit symmetry to generate other half}
\State $X \gets concat(X,X)$
\State $Y \gets concat(Y,Y)$
\State $Y \gets concat(Y,-Y)$
\State \Return (X,Y)
\EndProcedure

\Procedure{SampleSE}{$a,b,\epsilon,D$}
\State $\Theta(1) = 0$
\State $N = 1$
\While{$\Theta(N) < \frac{\pi}{2}$}
\State $\theta_{next} \gets UpdateTheta(\Theta(N),a,b,
\epsilon,D)$
\State $N = N + 1$
\State $\Theta(N) = \theta_{next}$
\EndWhile
\State $N = N + 1$
\State $\Theta(N) = \frac{\pi}{2}$
\While{$\Theta(N) > 0$}
\State $\theta_{next} \gets - UpdateTheta(\Theta(N),a,b,\epsilon,D)$
\State $N = N + 1$
\State $\Theta(N) = \theta_{next}$
\EndWhile
\State \Return $\Theta$
\EndProcedure

\Procedure{UpdateTheta}{$\theta,a,b,
\epsilon,D$}
\State $\theta_{\epsilon} \gets 0.01$
\If {$\theta \leq \theta_{\epsilon}$}
\State $\Delta_{\theta}(\theta) \gets (\frac{D}{b} - \theta^{\epsilon})^{\frac{1}{\epsilon}} - \theta$
\Else
\If {$\frac{\pi}{2} - \theta \leq \theta_{\epsilon}$}
\State $\theta_n \gets (\frac{D}{a} - (\frac{\pi}{2} - \theta)^{\epsilon})^{\frac{1}{\epsilon}} - (\frac{\pi}{2} - \theta)$
\Else
\State $\theta_n \gets \frac{\frac{D}{\epsilon} \cos(\theta) \sin(\theta)}{a^2 \cos^{2\epsilon}(\theta) \sin^4(\theta) + b^2 \sin^{2\epsilon}(\theta) \cos^4(\theta)}$ \Comment{Approximate arclength \refabeqp{arclength_approx_superellipse}}
\EndIf
\State $\Delta_{\theta}(\theta) \gets \theta_{n}$
\EndIf
\State \Return $\theta + \Delta_{\theta}(\theta)$
\EndProcedure
\end{algorithmic}
\end{algorithm}

\end{document}